\documentclass[sigconf, anonymous=False]{acmart}
\AtBeginDocument{%
  \providecommand\BibTeX{{%
    \normalfont B\kern-0.5em{\scshape i\kern-0.25em b}\kern-0.8em\TeX}}}

\copyrightyear{2021}
\acmYear{2021}
\setcopyright{acmcopyright}\acmConference[ICAIF'21]{2nd ACM International Conference on AI in Finance}{November 3--5, 2021}{Virtual Event, USA}
\acmBooktitle{2nd ACM International Conference on AI in Finance (ICAIF'21), November 3--5, 2021, Virtual Event, USA}
\acmPrice{15.00}
\acmDOI{10.1145/3490354.3494453}
\acmISBN{978-1-4503-9148-1/21/11}

\title{Financial misstatement detection: a realistic evaluation}
  
\begin{document}



\author{Elias Zavitsanos}
\email{izavits@iit.demokritos.gr}
\affiliation{%
  \institution{Inst. of Informatics \& Telecommunications, \\NCSR ``Demokritos"}
  \streetaddress{P.O. Box 60037}
  \city{}
%  \state{Aghia Paraskevi}
  \country{Greece}
%  \postcode{15310}
}

\author{Dimitris Mavroeidis}
\email{dmavroeidis@iit.demokritos.gr}
\affiliation{%
  \institution{Inst. of Informatics \& Telecommunications, \\NCSR ``Demokritos"}
  \streetaddress{P.O. Box 60037}
  \city{}
%  \state{Aghia Paraskevi}
  \country{Greece}
%  \postcode{15310}
}

\author{Konstantinos Bougiatiotis}
\email{bogas.ko@iit.demokritos.gr}
\affiliation{%
  \institution{Inst. of Informatics \& Telecommunications, \\NCSR ``Demokritos"}
  \streetaddress{P.O. Box 60037}
  \city{}
%  \state{Aghia Paraskevi}
  \country{Greece}
%  \postcode{15310}
}

\author{Eirini Spyropoulou}
\email{espyropoulou@iit.demokritos.gr}
\affiliation{%
  \institution{Inst. of Informatics \& Telecommunications, \\NCSR ``Demokritos"}
  \streetaddress{P.O. Box 60037}
  \city{}
%  \state{Aghia Paraskevi}
  \country{Greece}
%  \postcode{15310}
}

\author{Lefteris Loukas}
\email{eloukas@iit.demokritos.gr}
\affiliation{%
  \institution{Inst. of Informatics \& Telecommunications, \\NCSR ``Demokritos"}
  \streetaddress{P.O. Box 60037}
  \city{}
%  \state{Aghia Paraskevi}
  \country{Greece}
%  \postcode{15310}
}

\author{Georgios Paliouras}
\email{paliourg@iit.demokritos.gr}
\affiliation{%
  \institution{Inst. of Informatics \& Telecommunications, \\NCSR ``Demokritos"}
  \streetaddress{P.O. Box 60037}
  \city{}
%  \state{Aghia Paraskevi}
  \country{Greece}
%  \postcode{15310}
}

\renewcommand{\shortauthors}{Zavitsanos et al.}

\begin{abstract}
  In this work, we examine the evaluation process for the task of detecting financial reports with a high risk of containing a misstatement. This task is often referred to, in the literature, as ``misstatement detection in financial reports''. We provide an extensive review of the related literature. We propose a new, realistic evaluation framework for the task which, unlike a large part of the previous work: (a) focuses on the misstatement class and its rarity, (b) considers the dimension of time when splitting data into training and test and (c) considers the fact that misstatements can take a long time to detect. Most importantly, we show that the evaluation process significantly affects system performance, and we analyze the performance of different models and feature types in the new realistic framework.
\end{abstract}

\begin{CCSXML}
<ccs2012>
<concept>
<concept_id>10002951.10003317.10003359</concept_id>
<concept_desc>Information systems~Evaluation of retrieval results</concept_desc>
<concept_significance>500</concept_significance>
</concept>
<concept>
<concept_id>10002951.10003317.10003347.10003356</concept_id>
<concept_desc>Information systems~Clustering and classification</concept_desc>
<concept_significance>500</concept_significance>
</concept>
<concept>
<concept_id>10010147.10010257.10010258.10010259.10003268</concept_id>
<concept_desc>Computing methodologies~Ranking</concept_desc>
<concept_significance>500</concept_significance>
</concept>
</ccs2012>
\end{CCSXML}

\ccsdesc[500]{Information systems~Evaluation of retrieval results}
\ccsdesc[500]{Information systems~Clustering and classification}
\ccsdesc[500]{Computing methodologies~Ranking}

\keywords{misstatement detection, financial reports, 10-K, risk assessment}

\maketitle

\section{Introduction}
\label{sec:introduction}
Financial auditing is a long and time-consuming process that aims to ensure that everything reported in the annual financial report of a company is correct, i.e., that there are no misstatements. To do this, auditors check the financial statements of the report and compare them against the company's financial transactions. The auditors base their investigation on an initial assessment of risk factors associated with each company. The more risk factors they know about at the beginning of an audit, the more complete the assessment usually is.

The annual reports of publicly traded companies constitute open information, while known historical misstatements are recorded in widely available databases. The academic literature of accounting and auditing has used such historical data to train Machine Learning (ML) models that give a probability that a financial report contains a misstatement, i.e., the financial report contains differences between the actual financial statement items and those required by applicable accounting standards. This task is often referred to as "misstatement detection". Such models can only provide an indication of misstatement, which may constitute one of the risk factors that auditors consider.

The majority of the existing work focuses on financial features as input to the ML models~\cite{Hennes2008,Kirkos2007,Kotsiantis2006,Bai2008,Deng2009,Ravisankar2011}. Some efforts also consider linguistic features from the text segments of the annual reports, specifically from the Management Discussion and Analysis (MD\&A) section~\cite{Humpherys2011,Goel2012,Goel2016}. However, there is no standardized dataset and evaluation for this problem, making the results of different methods and input features in existing work not directly comparable.

The problem is handled as a binary classification task considering the annual reports that contain a misstatement as the positive class and the rest as the negative. There is a high imbalance between the two classes because reports containing misstatements are a small fraction of the data. Most related work uses standard classification evaluation measures such as accuracy, which treats the two classes equally and uses under-sampling of the negative class to create a balanced dataset. This leads to an unrealistic evaluation scenario that does not provide a reasonable estimate of the performance of the ML models in the real world. 

In addition, most related work splits the data into training and test at random, ignoring the chronological order of the reports. Time is essential when applying ML to this problem. First, the temporal distance between training and testing is important since the business environment changes, and the ML models may need to be updated. Additionally, the time at which misstatements get detected is equally important. They often go undetected for many years after submitting a report, and when they get noticed, they usually affect a series of reports by the company.

This paper proposes a realistic evaluation framework for misstatement detection and presents the first systematic study of different types of input and labeling for the task. We compile two main datasets, including text and financial features, that are among the largest used. Specifically, Section~\ref{sec:existing_work} provides a review of the related work focusing on multiple aspects. Section~\ref{sec:methodology} describes our methodology to derive the datasets for our experimental study, as well as the methodology we follow for setting up the experiments in a realistic evaluation framework. Finally, Section~\ref{sec:experimentation} discusses the experimental results and Section~\ref{sec:conclusions} concludes the paper.

\section{Related work}
\label{sec:existing_work}
In this section, we present a critical review of existing work on misstatement detection in terms of (a) the evaluation frameworks, (b) the input features, (c) the labeling sources, and (d) the machine learning methodologies.

\begin{table*}
  \caption{Related work using financial statement data.}
  \label{tab:related_work}
  \begin{tabular}{ccccc}
    \toprule
    Reference & Dataset size & Features & Evaluation & ML Method\\
    \midrule
    Hennes et al~\cite{Hennes2008} & 630 samples & Financial & Turnover measure & LR\\
    Kirkos et al~\cite{Kirkos2007} & Pos: 38, Neg: 38 & Financial & Accuracy & DT, NN, BBN\\
    Kotsiantis et al~\cite{Kotsiantis2006} & 164 firms, 41 pos & Financial & Accuracy & DT, NN, BBN, SVM, KNN\\ 
    Bai et al~\cite{Bai2008} & 148 firms, 24 pos & Financial & Accuracy & CART \\
    Deng et al~\cite{Deng2009} & 100 samples & Financial & Accuracy & SOM, K-means\\
    Ravisankar et al~\cite{Ravisankar2011} & 202 firms, 101 pos & Financial & Accuracy, AUC, Recall & NN, SVM, LR, GA\\
    Hoogs et al~\cite{Hoogs2007} & 390 firms, 51 pos & Financial & Accuracy & GA\\
    Kiehl et al~\cite{Kiehl2005} & 27 firms, 10 pos & Financial & TP rate & GA\\
    Lin et al~\cite{Lin2015} & 576 firms, 129 pos & Financial & Accuracy & LR, NN  \\
    Humpherys et al~\cite{Humpherys2011} & 202 firms, 101 pos & Text & Accuracy & DT, NB\\
    Goel et al~\cite{Goel2012} & 7100 samples, 405 pos & Text & p-value & Chi-square \\
    Glancy et al~\cite{Glancy2011} & 100 samples & Text & Accuracy & Text mining\\
    Hajek et al~\cite{Hajek2017} & 622 samples, 311 pos & Text \& fin. & Accuracy, AUC, TP, TN & BBN, Ensemble\\
    Kim et al~\cite{Kim2016} & 3000 samples, 788 pos & Financial & Accuracy, G-mean & LR, SVM, BN\\
    Dutta et al~\cite{Dutta2017} & 64000 samples, 788 pos & Financial & Accuracy, AUC, TNR & DT, NN, SVM, BBN\\
    Dechow et al~\cite{Dechow2011} & 2190 samples, 896 pos & Financial & Accuracy, F-score & LR\\ 
    Green et al~\cite{Green1997} & Pos: 86, Neg: 95 & Financial, Accounting & Accuracy, Recall & NN\\ 
    Feroz et al.~\cite{Feroz2000} & 132 firms & Financial, Operational & Accuracy, Error rate & LR\\
    Abbasi et al.~\cite{Abbasi2012} & Pos: 815, Neg: 8191 & Financial & AUC, Precision, Recall & Ensemble\\
    Spathis et al~\cite{Spathis2002} & Pos: 38, Neg: 38 & Financial & Accuracy, Recall & LR\\
    Kaminski et al~\cite{Kaminski2004} & Pos: 79, Neg: 79 & Financial & Accuracy & Discr. Analysis\\
    Cecchini et al~\cite{Cecchini2010} & Pos: 132, Neg: 3187 & Financial & Accuracy, Recall & SVM\\
    Perols~\cite{Perols2011} & Pos: 51, Neg: 15934 & Financial & AUC & LR, SVM, NN, DT, Bagging\\  
    Purda et al~\cite{Purda2015} & Pos: 1130, Neg: 4895 & Text & Accuracy, Recall & Text mining\\
    Craja et al.~\cite{Craja2020} & Pos: 201, Neg: 963 & Text and Financial & Acc., AUC, Sens/vity, Spec/city & HAN\\
    \bottomrule
  \end{tabular}
\end{table*}

\subsection{Evaluation framework}
One of the main characteristics of misstatement detection is the rarity of misstatements compared to the total number of financial reports. Although there is such a severe class imbalance, it is commonly assumed that the a-priori probability of a misstatement is equal to that of an accurate financial report. Much work in the literature adopts this assumption (\cite{Kirkos2007, Deng2009, Ravisankar2011, Kiehl2005, Humpherys2011, Glancy2011, Hajek2017, Green1997, Spathis2002, Feroz2000, Kaminski2004}). 

A commonly followed technique for creating such balanced datasets,
is -given a set of positive examples- to select an equal number of negative cases from the same year and industry. The rarity of positive examples usually leads to small datasets comprising only a few dozens of examples for each class (\cite{Kirkos2007, Kiehl2005, Spathis2002, Kaminski2004, Green1997}) as shown in Table~\ref{tab:related_work}. If such models were deployed in the real world, where the class distribution is heavily imbalanced, the performance would be significantly different from that reported based on a balanced test set.

On the other hand, some efforts take a more realistic approach and consider the class imbalance of the domain to some degree, usually with a ratio of positives to negatives between 1:3 and 1:6 (\cite{Kotsiantis2006, Kim2016, Lin2015,Craja2020, Bai2008, Hoogs2007}). The majority of these approaches use accuracy to evaluate results, which treats the classes equally, and thus, the conclusions may be misleading. However, the authors in~\cite{Craja2020} also report results using ROC-AUC (area under the receiver operating characteristic curve), sensitivity and specificity, focusing on the positive class. 

Methods that consider an even stronger class imbalance (\cite{Goel2016, Dutta2017, Dechow2011, Abbasi2012, Cecchini2010, Purda2015, Perols2011, Bao2020}), also evaluate in terms of AUC (\cite{Dutta2017, Abbasi2012, Perols2011}), true negative rate (TNR)~\cite{Dutta2017}, Precision, and Recall (\cite{Abbasi2012, Cecchini2010, Purda2015}), as shown in Table~\ref{tab:related_work}. 

Most methods reviewed so far evaluate misstatement detection as a binary classification task. Many of those use random splits and cross-validation for training and testing without considering the dimension of time. This is very different from the real-world setting, where predictions need to happen on future financial reports. 

The recent work in \cite{Bao2020} considers these issues. It suggests that misstatement detection should be evaluated as an information retrieval task where results are ranked based on the predicted probability for containing a misstatement. The authors have used ROC-AUC, Normalized Discounted Cumulative Gain (NDCG), and Precision at a small percentage of the test set to evaluate various ML methods for this task. They have also considered the dimension of time to maintain the chronological order of the data when taking training and testing splits. However, their evaluation framework partially addresses the fact that misstatements usually take a long time to get detected. 

In our approach, we consider the severe class imbalance and avoid measures like accuracy. We focus on the positive class, and we evaluate the classifiers as rankers. Finally, we consider the dimension of time in the way we train and test and the time for a misstatement to get detected in the real world.

\subsection{Input features}
In terms of input features, there are mainly two commonly used types: financial and textual ones. Financial features may include indices and amounts found in the financial statements and ratios that are calculated from the numbers of the financial statements. Those features usually come from proprietary databases like COMPUSTAT\footnote{\url{https://www.marketplace.spglobal.com/en/datasets/compustat-fundamentals-(8)}} or are extracted from the annual statements themselves. On the other hand, textual features consider the tokens that appear in the text or other linguistic information. 

The features that are commonly used to train the classifiers are financial (\cite{Kirkos2007,Kotsiantis2006,Bai2008,Deng2009,Ravisankar2011,Hoogs2007,Kiehl2005,Chai2006,Liou2008}) as shown in Table~\ref{tab:related_work}, with the work in~\cite{Dechow2011} and~\cite{Cecchini2010} being among the most popular approaches used in the accounting literature. Both works provide a comprehensive analysis of financial indices, variables, and ratios. They also provide insights into the usefulness of different features in detecting misstatements that have inspired many other misstatement detection approaches. Other features may concern operational characteristics, such as the CEO's, CFO's, and previous auditor's turnover (\cite{Feroz2000}, \cite{Abbasi2012}), which are usually combined with financial features in this task.

There are a few approaches to misstatement detection that go beyond financial features. Most work in this category has focused on the text of the MD\&A section of financial reports. This section provides qualitative information and reflects the management's opinion and analysis of the reported financial results. Based on this information, the work in \cite{Purda2015} selects the best 200 words from a corpus of MD\&A sections to use as features for classification. Those words are assumed to be good predictors for misstatements, while in \cite{Humpherys2011} the emphasis is on the use of ``pleasant words", which are assumed to be used more commonly in fraudulent disclosures. Typically, these linguistic features come from pre-determined lists of words that have been associated with intentional misstatements. Additional features that have been used in the literature include readability measures, such as the average length of words and sentences, lexical diversity, and sentence complexity \cite{Laughran2011}. Similarly, the authors in \cite{Goel2012, Goel2016} use word lists that assign sentiment polarity (positive, negative, and neutral) to words. Finally, the work in \cite{Fissette2017} uses word tokens with their corresponding TF-IDF values to train classifiers and predict misstatements.

Finally, there are a few cases that combine financial and textual features to train the classifiers, such as the work in \cite{Hajek2017} which has shown that it is possible to improve the predictive performance of the learned model by combining the two types of features. Finally, a more recent approach \cite{Craja2020} has combined the two types of features by concatenating the neural encoding of the documents with the financial features before the final classification layer. All these approaches assume that text contains indications of a high probability of misstatement.

In this work, we compile datasets containing textual and financial features to examine whether one type of feature prevails over the other under our evaluation framework.

\subsection{Labeling sources}
Various databases have been used as sources to label the training examples in the datasets throughout the related literature. In datasets based on US companies, the most commonly used are the US Government Accountability Office\footnote{\url{https://www.gao.gov/}} (GAO), the US Securities and Exchange Commission (SEC) Accounting and Auditing Enforcement Releases\footnote{\url{https://sites.google.com/usc.edu/aaerdataset/buy-the-data?authuser=0}} (AAER), and the Audit Analytics\footnote{\url{https://www.auditanalytics.com}} (AA) database. 

GAO is a compilation of restatement announcements that cover misstatements associated mainly with financial misrepresentation. AAER is a designation assigned by the SEC to administrative proceedings or litigation releases and covers intentional misstatements. Finally, AA tracks financial restatements in public filings from EDGAR\footnote{\url{https://www.sec.goc/edgar.shtml}}, the SEC's Electronic Data Gathering, Analysis, and Retrieval system. A comparison of these databases appears in \cite{Karpoff2012}.

AAER labels have been mainly used in cases where the focus is on identifying intentional misstatements (\cite{Green1997, Fissette2017, Feroz2000, Lin2015, Bao2020, Craja2020, Dechow2011, Hajek2017, Perols2011, Cecchini2010, Abbasi2012}). On the other hand, GAO is less widely used \cite{Kim2016}. AA, although it contains a comprehensive set of various types of restatement cases, with more information than the other databases, it is also seldomly used \cite{Dutta2017}. In cases where data samples come from countries other than the US, other labeling mechanisms have been used, such as the Athens Stock Exchange and taxation databases in \cite{Kotsiantis2006, Kirkos2007}, the Chinese Stock Exchange in \cite{Ravisankar2011}, the Istanbul Stock Exchange in \cite{Ata2009}, and more.

A significant factor that prevents the direct comparison of different methods is the use of different labeling mechanisms. In this study, we use two different mechanisms to compile our datasets, AAER and AA, aiming to explore the effect of using different types of labels on the performance of the classifiers.

\subsection{Machine learning methods}
The majority of the related work is mainly concerned with selecting input features. In terms of ML methods, we mainly observe the application of many traditional algorithms (see Table~\ref{tab:related_work}), with many approaches examining the performance of more than one classifier \cite{Sharma2012}. In this direction, the authors in \cite{Kirkos2007} use decision trees (DT), neural networks (NN), and Bayesian belief networks (BBN), with the latter achieving the best results, while the work in \cite{Kotsiantis2006} also examines the performance of Support Vector Machines (SVM) and K-Nearest Neighbours (KNN). SVMs were also used in \cite{Cecchini2010}.

In another study~\cite{Liou2008}, logistic regression (LR), neural networks, and classification trees were considered, with logistic regression showing a better performance than the others, while in a more recent work \cite{Ravisankar2011} neural networks achieved the best results. In contrast, the authors in \cite{Bai2008} achieve better results with regression trees (CART). In general, we mainly find comparative studies that explore the performance of various ML methods (\cite{Ata2009, Lin2015, Perols2011}), that usually result in different conclusions because (i) the datasets are different, (ii) the features are different, and (iii) there is no standard common evaluation framework that considers the particular characteristics of the domain. 

When using text features, few studies apply natural language processing (NLP) techniques to represent the entire textual content of the financial reports~\cite{Fissette2017}. Furthermore, to our knowledge, there is only a single method that has used deep learning models or deep neural representations for text analysis in financial statements. The authors in ~\cite{Craja2020} use a hierarchical attention network to model texts, and they concatenate the document encoding with the financial features before the final classification layer. This paper focuses on comparing types of features and labeling sources in a realistic evaluation framework. Thus we only experiment with traditional ML methods.

\section{Methodology}
\label{sec:methodology}
In this section, we describe the methodology for compiling the datasets, and we present our proposed evaluation framework. 

\subsection{Input curation}
\label{sec:data}

The datasets used in this paper comprise financial and textual information from the annual reports of public US companies. The annual reports are filed in a 10-K form\footnote{\url{https://www.sec.gov/files/form10-k.pdf}}, as described by the SEC. These filings provide a snapshot of a company's financial profile, including quantitative and qualitative information about the company.

\begin{table*}
  \caption{Financial feature set from~\cite{Bao2020} and~\cite{Dechow2011}.}
  \label{tab:fin_features}
  \begin{tabular}{ccl}
    \toprule
%    \midrule
 Current assets, total (act) & Account payable (ap) & Assets, total (at) \\
 Common/ordinary equity, total (ceq) &  Cash \& short investments (che) & Cost of goods sold (cogs)\\
 Common shares outstanding (csho) & Debt in current liabilities, total (dlc) & Inventories, total (invt)\\
 Long-term debt issuance (dltis) & Long-term debt, total (dltt) & Net income (ni)\\
 Property \& equipment, total (ppegt) & Depreciation \& amortization (dp) & Receivables, total (rect)\\
 Income bf extraordinary items (ib) & Investment and advances, other (ivao) & Retained earnings (re)\\
 Short-term inv/ments, total (ivst) & Current liabilities, total (lct) & Liabilities, total (lt)\\
 Preferred stock (capital), total (pstk) & Sales/turnover (sale) & Sale of common/preferred stock (sstk)\\
 Income taxes payable (txp) & Income taxes, total (txt) & Interest and expense, total (xint)\\
 Price close, annual, fiscal (prcc\_f) & WC accruals (dch\_wc) & RSST accruals (ch\_rsst)\\
 Change in receivables (dch\_rec) & Change in inventory (dch\_inv) & Change in cash sales (ch\_cs)\\
 Soft assets percentage (soft\_assets) & Change in cash margin (ch\_cm) & Change in return on assets (ch\_roa) \\
 actual issuance (issue) & Book to market (bm) & Depreciation index (dpi)\\
 Retained earnings over assets (reoa) & Earnings before interest (EBIT) & Cash Flow Earnings Difference (ch\_fcf)\\
\bottomrule
  \end{tabular}
\end{table*}

Specifically, we used the financial information of the public dataset from Bao et al.~\cite{Bao2020}, which contains a list of public US companies with financial indices that come from consolidated financial statements. Each row in this dataset corresponds to a financial report of a specific year. For each such example, 28 economic indices are provided that reflect items from the financial statements. We enriched this dataset with 14 additional features that we calculated based on Dechow et al.~\cite{Dechow2011}, which is regarded as the most comprehensive prediction model. These extra features reflect changes and ratios of the financial indices. Table~\ref{tab:fin_features} presents the list of financial features used in our dataset. This dataset contains 100K examples that span between the years 1991 and 2008. We kept the data up to 2008 to experiment in the same years as in~\cite{Bao2020}. Additionally, access to resources with financial information for recent years is not trivial.

In addition to the financial indices, our dataset includes qualitative information derived from the financial reports' text. We downloaded the 10-K annual filings from EDGAR using the edgar-crawler in~\cite{loukas-2021-edgar-corpus}. We extracted the text from the MD\&A section and performed basic segmentation and cleaning to remove noise, HTML tags, symbols, inline attachments, and non-textual objects. Finally, we aligned the textual information with the financial features based on the period of report for each example.

\subsection{Dataset creation}\label{subsec:data_creation}

\begin{table*}
  \caption{Distribution of positive/negative examples over the years.}
  \label{tab:pos_dist}
  \begin{tabular}{cccccccccccc}
    \toprule
    Source of labels/Year & 1998 & 1999 & 2000 & 2001 & 2002	& 2003 & 2004 & 2005 & 2006 & 2007 & 2008\\
    \midrule
    \textbf{AAER} & 27/3094 & 43/3141 & 52/3035 & 53/2935 & 57/2845 & 54/2789 & 44/2822 & 32/2747 & 21/2811 & 21/2878 & 17/2925\\
    \textbf{AA} & 11/3110 & 21/3163 & 30/3057 & 51/2937 & 55/2847 & 41/2802 & 51/2815 & 42/2737 & 38/2794 & 50/2849 & 67/2875\\
    \midrule
    \textbf{Total samples} & 3121 & 3184 & 3087 & 2988 & 2902 & 2843 & 2866 & 2779 & 2832 & 2899 & 2942\\
\bottomrule
  \end{tabular}
\end{table*}

Each example (i.e., report) is also associated with a label specifying whether or not it was found to contain a misstatement. Examples that contain a misstatement are marked as positives and the rest as negatives. We used two different labeling sources. The first one consists of labels from AAER, which were already available in~\cite{Bao2020}. Additionally, we used the labels from AA.
Although the two sets of labels are different, since AAER has mainly intentional misstatements, and AA has misstatements that may have happened due to accounting errors, they contain almost the same number of positive examples. Table~\ref{tab:pos_dist} shows the number of positive and negative examples per year.

Based on the two sources of labels and the two input types, we create four datasets: (a) \textit{AAER-Fin}, containing financial input and labeled based on the AAER database, (b) \textit{AAER-Txt}, containing text input and labeled based on the AAER database, (c) \textit{AA-Fin}, containing financial input and labeled based on the AA database, (d) \textit{AA-Txt}, containing text input and labeled based on the AA database. All datasets include 32443 samples from 1998 to 2008, with the positives accounting for only about 1\% of the total instances. 

In the real world, we may not be aware of a company's misstatements until two or three years after they happen~\cite{Dyck2010}. In addition, when a misstatement gets detected, it usually affects also the reports of previous years. Ignoring this fact may have a significant influence on the observed performance. Interestingly, the AA database contains the restatement date for every misstatement. This is essentially the date that the report was resubmitted with corrections, which is close to the date that the misstatement was detected. We derive two additional datasets, namely \textit{AA-Hard-Txt} and \textit{AA-Hard-Fin}, which contain the re-statement date for every positive label. We use this date in our experiments to create an even more realistic evaluation scenario.

More information about the dataset creation can be found in Appendix A.

\subsection{Evaluation framework}
\label{sec:meth:experimental_setting}

In a real-world application, a misstatement detection system will be trained only with past examples since these examples will be available and is expected to predict misstatements on annual reports of the latest fiscal year.
This is translated to using examples up to year \emph{Y-1} for training and parameter tuning and use year \emph{Y} only for testing. It is not realistic to shuffle the data to train and test on random sets, i.e., mixing into the training set future samples that may ``leak" financial changes that should have been unknown to that point.
Also, given the imbalanced class distribution, we expect the majority of reports for year \emph{Y} to belong to the negative class. We consider it essential to maintain this realistic class distribution for tuning and especially testing. 

\begin{figure}[h]
  \centering
  \includegraphics[width=\linewidth]{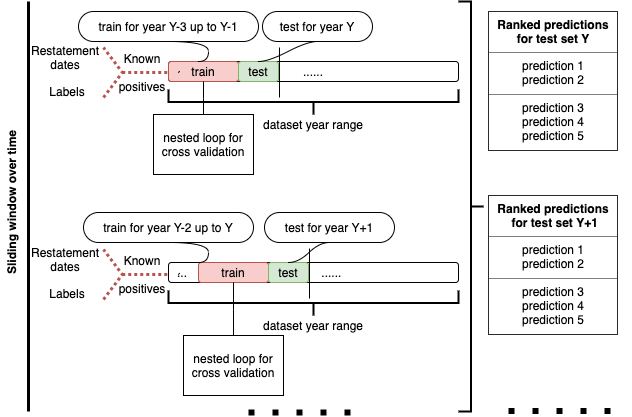}
  \caption{The proposed evaluation framework.}
  \label{fig:evaluation}
\end{figure}

With that in mind, we set up the experiments following a chronological order to create the folds. As shown in Figure~\ref{fig:evaluation}, we use a sliding window of three years for training, and we always test on the succeeding year. For example, when testing in 2003, we use 2000-2002 for training and tuning. Then, we slide the training window to 2001-2003 and test on 2004, and so on. We use a nested loop applying 5-fold cross-validation for tuning inside every training window. We assess the performance of different models on the same held-out test sets to compare results. 

We further extend our evaluation to simulate the fact that misstatements take time until they get detected in the real world. In practice, this means that, if, for example, we were to predict misstatements for 2003, we would probably not have all the positive examples identified in the corresponding training set of 2000-2002. 
Using the restatement date, which is available for every positive example in the AA-hard datasets, we flip every positive label to negative for every \emph{training fold} if its restatement date is later than the test year for that fold. Essentially, this affects all three years that are used for training. Of course, labels of future filings (test years) are not flipped since they constitute the ground truth.

Finally, we use and evaluate the classifiers as rankers because using these models as risk factors for misstatement in a financial report requires the association of every report with a score, i.e., the classifier's confidence. Also, given the limited time capacity of human auditors, we are interested in classifiers that perform well in the ranking measures rather than just the classification measures to safely choose to see as many of the high-risk reports as they can.

In all of our experiments, we report results on R-precision, which is a ranking evaluation measure. It uses the number of positive examples (\emph{R}) in the test set as a cutoff to measure the actual precision and recall. Thus, R-precision reflects the model's ability to propose correct predictions (\emph{r}) when we give the system as many opportunities as the number of positive examples, reflecting at the same time precision at \emph{R} and recall. R-precision is defined in Equation~\ref{eq:rprecision}.

\begin{equation}
\label{eq:rprecision}
  Rprecision = \frac{r}{R}
\end{equation}

\section{Experiments}
\label{sec:experimentation}
We divide this section into three parts. First, we describe the data pre-processing steps. Then, we explore the comparison between using text and financial features as input and the effect of using different labeling sources (AAER, AA). Then, we experiment by simulating the effect of time delay in detecting misstatements in the real world, using the AA-Hard dataset, aiming to explore the actual performance of the models under this realistic setting. In terms of ML methods, we choose to experiment with SVM and logistic regression (LR), that are also used in the related literature. We use their weighted variants to address the issue of class imbalance. We tuned the hyper-parameters \emph{C} and \emph{kernel} for SVM and \emph{C} for Logistic Regression (see Appendix~\ref{AppendixB} for details). \emph{Negative log loss} was used as the evaluation metric for hyper-parameter optimization. Experiments were conducted using the scikit-learn machine learning library~\cite{scikit-learn}.

% Before settling to R-Precision as our measure of choice, we experimented with several other seemingly suitable measures such as ROC-AUC, Precision@k and F1-macro. None of them exhibited the required characteristics for the task. ROC-AUC, for example, presented very high variability compared to other metrics. F1-macro was very stable, but is not a ranker, which is a requirement in our proposed framework.

\subsection{Data pre-processing and features}
As explained in Section~\ref{sec:data}, we included both financial and text features in our datasets. 
Regarding the financial features, besides calculating the additional ratios, we regularize them using standard scaling.

Regarding text, most related work uses word counts vectors (bag of n-grams), or linguistic features, while only one~\cite{Craja2020} uses word embeddings.
Since our primary objective is to present a systematic study with a realistic evaluation framework for misstatement detection, rather than achieving state-of-the-art results, we opted for the simple representation of ``bag-of-ngrams". The text extracted from the financial reports was converted to token vectors. We employed the usual pre-processing steps of sentence splitting, word tokenization, and lowercasing.
In addition, we performed pre-processing tailored to financial documents; we replaced all amounts with the ``\textit{amount\_replaced}" token, all dates with the ``\textit{date\_replaced}" token, and the remaining numbers with the ``\textit{number\_replaced}" token. This process significantly reduced the vocabulary size, leading to less sparse token vectors and grouping of similar information. Finally, we calculated the normalized TF-IDF value per financial report for each token and used it in the corresponding vector. We consider tokens to be uni-grams (single words) and bi-grams (sequences of two words), capturing local context. The resulting vocabularies contain an average of 3.8M tokens per dataset.

 Additionally, we combined textual and financial features, adjusting their weights for optimal performance. When using combined features, textual features were the dominant predictors every time. To reduce the dimensionality of the feature space and improve predictive capability, we also employed methods to select the most important features. We used two popular methods: Univariate features selection and meta selection using a linear SVM estimator. None of these methods worked as expected, reducing the predictive capabilities of all models.

\subsection{The effect of different labels}
\label{subsec:different_labels}
\begin{figure}[h]
  \centering
  \includegraphics[width=0.85\linewidth]{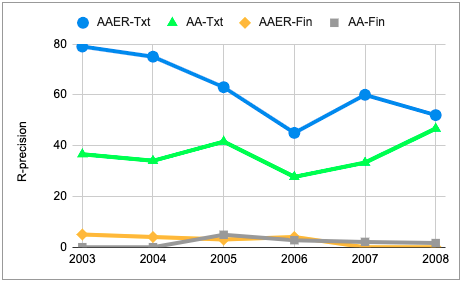}
  \caption{Performance of SVM in AAER and AA datasets.}
  \label{fig:bao_vs_aa}
\end{figure}

Here, we explore differences in performance between models trained using the two different labeling sources (AA and AAER). Figure~\ref{fig:bao_vs_aa} illustrates the performance of SVM\footnote{For clarity reasons, we do not show results with LR since the performance is almost identical in all cases.}, in terms of R-precision, using text or financial features (\textit{AAER-Txt}, \textit{AA-Txt}, and \textit{AAER-Fin}, \textit{AA-Fin} respectively). SVMs using text-based features significantly outperform financial features.

The AAER labels capture irregularities and intentional misstatements, making the task easier than AA labels, as classes are more distinguishable than those in AA. We assume that AA positive samples are harder to identify because they may include unintentional accounting errors that are not easily reflected in the input features. The superior performance of the \textit{AAER-Txt} model against the \textit{AA-Txt} confirms this assumption.

\begin{figure}[h]
  \centering
  \includegraphics[width=\linewidth]{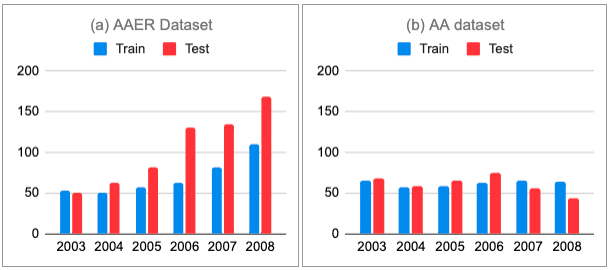}
  \caption{Ratio of negative to positive examples in the training and test sets for each evaluation year. (a) Dataset with AAER labels. (b) Dataset with AA labels.}
  \label{fig:neg_pos_ratio}
\end{figure}

Regarding \textit{AAER-Txt}, the sudden drop in 2006 and the lower performance afterward is due to a change in data composition for these years. In particular, the number of positive examples is much lower (Table~\ref{tab:pos_dist}) in these years, while the number of negative examples remains about the same. Thus, the negative-to-positive ratio inside the training sets and the test sets increases, as shown in Figure~\ref{fig:neg_pos_ratio}(a). On the other hand, the corresponding fluctuation in the AA dataset is much smaller, as shown in Figure~\ref{fig:neg_pos_ratio}(b). However, the very few positives in 2006 also affected the performance of \textit{AA-Txt} in that year. The changes in data composition are not reflected in \textit{AAER-Fin} and \textit{AA-Fin}, where poor performance in both datasets imply that financial features are inferior predictors to the textual ones.

The experimental setting in this section is based on an ideal scenario that assumes misstatements are detected as soon as they occur. The following section explores how the performance is affected when we assume that positive examples are not detected immediately.

\subsection{Experimentation in the most realistic scenario}
As already stated, misstatements get detected on average two to three years after they occur. When they get detected, they usually affect the reports of a company for a number of consecutive years. Ignoring this fact may significantly influence the observed performance, especially when considering the text as input. In particular, we expect the learned models to over-fit on ``serial'' misstatements. In other words, if a company exhibits a misstatement in the training data, many words or phrases from that report that are unique to this company (i.e., the name of the company or its products) could be used to label any later statement as a misstatement as well. Thus, the text model would essentially learn to label the statements based mainly on past behavior of specific companies. Indeed, this could explain the higher performance of models on the \textit{AAER-txt} and \textit{AA-Txt} datasets presented in Section~\ref{subsec:different_labels}. To test this hypothesis, we only used the company name as feature when training the SVM models. We observed almost identical performance to using text from the financial reports, hence confirming our suspicion of over-fitting on the identity of specific companies.

Moving towards the real-world case, where misstatements are detected with delay, we use the proposed evaluation framework of Section~\ref{sec:meth:experimental_setting} to compare the performance of the SVM in the case of keeping all the positive labels versus flipping some of them to negative, based on their restatement date. The former case is represented with experiments on the \textit{AA-Txt} and \textit{AA-Fin} datasets and the later with experiments on the \textit{AA-Hard-Txt} and \textit{AA-Hard-Fin} datasets.

\begin{figure}[h]
  \centering
    \includegraphics[width=0.85\linewidth]{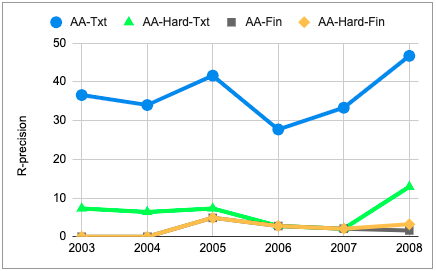}
  \caption{Performance of SVM in AA and AA-Hard datasets.}
  \label{fig:use_restatement_date}
\end{figure}

\begin{figure}[h]
  \centering
  \includegraphics[width=0.9\linewidth]{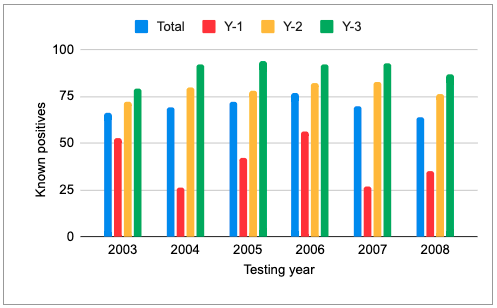}
  \caption{The percentage of the positive examples (misstatements) that are known per training year. \emph{Y-1}, \emph{Y-2}, and \emph{Y-3} correspond to the three years before the testing year.}
  \label{fig:pos_percent_each_year}
\end{figure}

As shown in Figure~\ref{fig:use_restatement_date}, there is a significant drop in the performance of the SVM using text input when considering the real-world case (AA-Hard datasets). It is worth noting that almost 40\% of the positive training examples have flipped labels and appear as negative samples, as Figure~\ref{fig:pos_percent_each_year} shows. As expected, this percentage is higher for the year that is immediately before the testing year. For example, to evaluate in 2003, based on the restatement information, we keep only about 50\% of the positive examples in \emph{Y-1} year, i.e., in 2002, we keep 70\% of the positive examples in the year 2001 (\emph{Y-2} year), and about 80\% of the positives for the year 2000 (\emph{Y-3} year).

As we move to more recent years, the effect of unknown misstatements gets worse. The percentage of known positive examples for year \emph{Y-1} further decreases. This year contains the most up-to-date information regarding the company's financial status and thus provides the more recent ``snapshot" of its trends and momentum. However, it misleads the training process due to many unknown positive labels that appear as negatives.

\begin{figure}[h]
  \centering
  \includegraphics[width=0.85\linewidth]{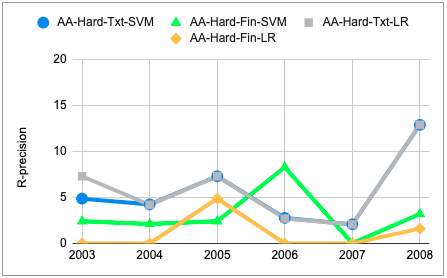}
  \caption{Performance of different models in the AA-hard dataset.}
  \label{fig:models_3y_train_no_gap}
\end{figure}

Overall, all tested models have similarly low performance in the \textit{AA-Hard} datasets with our proposed evaluation setting. We observe similar low performance using text, financial features, or a combination of the two. Figure~\ref{fig:models_3y_train_no_gap} shows a comparison between SVM and LR using the \textit{AA-Hard} datasets with our proposed evaluation setting, where we do not observe any significant differences in performance.

The most closely related work to our approach is that of Bao et al.~\cite{Bao2020}. Their proposed solution to handle undetected misstatements is to leave a two-year gap between training and testing, thus excluding the training years that may have the most considerable amount of misstatements. Our exploratory analysis confirmed that these two years have the largest amount of unknown positives (Fig.~\ref{fig:pos_percent_each_year}), but unknown positives exist in previous years as well. For reasons of completeness, we have also experimented with their approach. Our results, however, are not directly comparable to theirs due to the different datasets. Besides financial features, we also use text, and we also simulate the effect of time delay in misstatement detection in the \textit{AA-Hard} datasets. Thus we also have different labels. 

\begin{figure}[h]
  \centering
\includegraphics[width=0.85\linewidth]{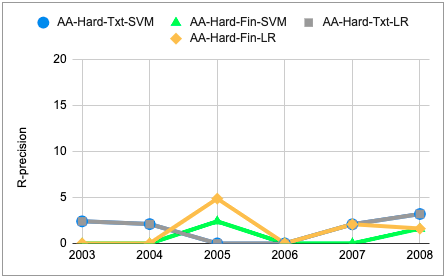}
  \caption{Performance of different models in the AA-hard dataset, leaving a 2-year gap between training and testing.}
  \label{fig:models_3y_train_2_gap}
\end{figure}

Figure~\ref{fig:models_3y_train_2_gap} shows that, if the last two years are ignored completely, performance further deteriorates.
This observation may indicate that it becomes harder to find consistent misstatement patterns when the time gap between training and testing increases. Intuitively, this may be caused by the fact that the financial landscape changes over time, and the financial features only capture a snapshot of the business for a short period. However, these insights need further investigation to be validated.

We have further experimented with one-year and two-year gaps between the training and the test sets, in combination with the use of three or five years for training, with no improvement in performance. The best results for \textit{AA-hard} are those shown in Figure~\ref{fig:models_3y_train_no_gap}. 

%Table~\ref{tab:overall_results} presents additional results, showing how recall increases when more examples are retrieved, i.e., 100, 150, 200 and 250, that correspond to 4\%, 6\%, 8\% and 10\% of the test set.

%\begin{table*}
%  \caption{Average results over all test sets.}
%  \label{tab:overall_results}
%  \begin{tabular}{ccccccl}
%    \toprule
%    Model & ROC\_AUC & R@100 & R@150 & R@200 & R@250\\
%    \midrule
%    LR-text & 0.526 & 0.075 & 0.103 & 0.134 & 0.156\\  
%    SVM-text & 0.537 & 0.075 & 0.091 & 0.105 & 0.117\\
%    LR-fin &  0.528 & 0.030 & 0.065 & 0.085 & 0.108\\
%    SVM-fin & 0.521 & 0.066 & 0.078 & 0.091 & 0.109\\
%\bottomrule
%  \end{tabular}
%\end{table*}

\section{Conclusions}
\label{sec:conclusions}
In this work, we studied the evaluation of machine learning methods for detecting financial reports with a high risk of containing a misstatement. We used and evaluated the trained classifiers as rankers, using an appropriate evaluation measure. 

The primary purpose of this work was to emphasize and illustrate the need for a realistic evaluation, which is often missing in the corresponding literature. We, therefore, proposed a realistic chronological evaluation framework, maintaining the severe class imbalance in the dataset and -at the same time- simulating the fact that the positives are not detected as soon as they happen. Through empirical evaluation, we showed the effect of having different datasets and input features, and we revealed that the task is inherently harder than suggested in the literature.

Based on the experimental observations, future plans include exploring neural text representations combined with deep learning models. At the same time, the unknown positives in the negative class suggest the use of Positive - Unlabeled (PU) learning methods.

%\begin{acks}
%\end{acks}

\bibliographystyle{ACM-Reference-Format}
\bibliography{main}

\appendix

\section{Datasets}\label{AppendixA}
This section provides additional details about the compilation of the datasets we used. We cannot distribute them publicly because parts of them are from proprietary sources.

We used the dataset from Bao et al.~\cite{Bao2020} to get the financial figures for each example (i.e., report). At the time this paper was written, these data were available on GitHub\footnote{https://github.com/JarFraud/FraudDetection}. We enriched each report with textual data as described in section~\ref{sec:data}, by using the edgar-crawler~\cite{loukas-2021-edgar-corpus} to download the 10-K annual filings from EDGAR. In order to align each report's financial data with the corresponding textual data, we matched the company key provided in the initial dataset of~\cite{Bao2020} with the Central Index Key (CIK) from EDGAR. COMPUSTAT provides this matching. At the end of this process, we had an enriched version of the Bao et al. dataset. Each row represents an annual report of a company with financial and textual information and a label that indicates whether it contains a misstatement or not. These labels are based on AAERs and are directly provided in the Bao et al. dataset.

The next step was to create a different version of the dataset using the AuditAnalytics as a second source for misstatement labels, as described in section~\ref{subsec:data_creation}. For the selected time period, we queried the misstatement database of AuditAnalytics for 10-K reports that contain a misstatement. The result of this query also includes the restatement date for each report that contains a misstatement. We use this information to simulate the effect of time delay in detecting misstatements in the real world in our realistic framework. 

Assume we have a report \emph{R} in the year 2001 that is labeled as a misstatement, and the corresponding restatement date was in 2003. When we start the training process, as described in section~\ref{sec:meth:experimental_setting}, we use the years 2000 up to 2002 for training and 2003 for testing. Thus, at the time of testing, the misstatement of report \emph{R} would not have been identified yet. Therefore we flip its label and use it as a non-misstatement in the training set (although a-posteriori, we know this was a misstatement). Next, as we move the sliding window for training, we use 2001 to 2003 for training and 2004 for testing. Thus, at the time of testing, we know that report \emph{R} contains a misstatement (since its restatement date is 2003). Therefore, in the training set that we have now, we use report \emph{R} as a positive example. We perform this process only for training to simulate having unknown (i.e., unidentified) positive examples in the training sets. We maintain the gold labels at the test sets at all times.

\section{Tuning of hyper-parameters}\label{AppendixB}
This section presents the hyper-parameter values that resulted from the tuning process for the strict yet realistic evaluation setting, using the AA-hard dataset. We tuned the following hyper-parameters in the following ranges:

\begin{itemize}
\item LR - C: \{ 0.01, 0.1, 0.5, 1.0, 2.5, 5.0, 10.0, 20.0  \} 
\item SVM - C: \{ 0.01, 0.1, 0.5, 1.0, 2.5, 5.0, 10.0, 20.0  \} 
\item SVM - kernel: \{ Linear, Polynomial, RBF \}
\end{itemize}

Recall from section~\ref{sec:meth:experimental_setting}, that training is performed based on a sliding window that trains models for each test year. Thus, hyper-parameter tuning is performed for each test year. Tables~\ref{tab:tune_text} and~\ref{tab:tune_fin} provide the best parameters found using text or financial features respectively.

\begin{table}[htb]
 \caption{Hyper-parameter values for text features.}
 \label{tab:tune_text}
 \begin{tabular}{lccc}
 \toprule
 & \multicolumn{1}{c}{LR} & \multicolumn{2}{c}{SVM} \\
 \midrule
Test year & C & kernel & C \\
\midrule
2003 & 20.0 & RBF & 20.0 \\
2004 & 20.0 & RBF & 20.0 \\
2005 & 20.0 & RBF & 20.0 \\
2006 & 20.0 & RBF & 20.0 \\
2007 & 20.0 & RBF & 20.0 \\
2008 & 20.0 & RBF & 20.0 \\
\bottomrule
\end{tabular}
\end{table}
\begin{table}[htb]

 \caption{Hyper-parameter values for financial features.}
 \label{tab:tune_fin}
 \begin{tabular}{lccc}
 \toprule
 & \multicolumn{1}{c}{LR} & \multicolumn{2}{c}{SVM} \\
 \midrule
Test year & C & kernel & C \\
\midrule
2003 & 20.0 & RBF & 20.0 \\
2004 & 20.0 & RBF & 20.0 \\
2005 & 20.0 & RBF & 5.0 \\
2006 & 20.0 & RBF & 20.0 \\
2007 & 20.0 & RBF & 20.0 \\
2008 & 20.0 & Polynomial & 20.0 \\
\hline
\end{tabular}
\end{table}

\end{document}